\documentclass[runningheads]{llncs}
\usepackage[T1]{fontenc}
\usepackage{graphicx}
\usepackage{hyperref}
\usepackage{color}

\def\bdq#1{
\lq\lq{#1}\rq\rq}
\usepackage{comment}
\usepackage{commath}
\usepackage{booktabs}
\usepackage{amsmath,amssymb,amsfonts}
\usepackage{multirow}
\usepackage{subcaption}
\usepackage{tabularx}
\usepackage{multicol}

\newcommand{\dop}[1]{\textcolor{black}{{#1}}}
\newcommand{\rev}[1]{\textcolor{black}{{#1}}}

\usepackage[normalem]{ulem}

\begin{document}
\title{Leveraging contextual events on structure-aware next activity prediction}
\titlerunning{Leveraging contextual events on structure-aware next activity prediction}
\author{
Alessandro Mele\inst{1}\orcidID{0009-0009-7852-0528}
\and Claudia Diamantini\inst{1}\orcidID{0000-0001-8143-7615}
\and Domenico Potena\inst{1}\orcidID{0000-0002-7067-5463}
}
\authorrunning{A. Mele et al.}
\institute{Department of Information Engineering, Polytechnic University of Marche, Ancona, Marche, Italy}
\maketitle

\begin{abstract}
Predictive process monitoring aims at forecasting various aspects of running processes. Among the different tasks, next activity prediction represents the most extensively investigated. However, only a limited number of existing approaches explicitly encode contextual information, i.e., the environmental conditions in which the process is executed, typically modeled through event log attributes or aggregated measures. In this paper, an approach based on the concept of Instance Graphs is introduced. To incorporate contextual process instances, several encoding strategies are proposed and evaluated by measuring their impact on prediction performance. For each encoding strategy, a set of prefix-Instance Graphs is generated and subsequently provided as input to a Graph Neural Network for the classification task. The proposed approach is evaluated on multiple real-world event logs, and the experimental results demonstrate that \rev{incorporating contextual process instances benefits prediction performance.}

\keywords{next activity prediction \and contextual events \and graph neural networks}
\end{abstract}

\section{Introduction}
\label{sec:introduction}
Predictive process monitoring (PPM) is a branch of process mining that leverages historical data to predict the evolution of ongoing process instances. In the last decade, the scientific community shifted from rule-based predictions to data-driven approaches, in which models are trained on historical data to automatically learn process behaviors \cite{DiFrancescomarino2022}. PPM tasks can be categorized in three fields, namely the predictions of categorical outcome values, measures of interest taking continuous values, and the sequences of future activities and related data payloads, i.e., next event predictions \cite{DiFrancescomarino2022}. To perform prediction, most of the existing approaches model process executions \rev{without considering that a process typically executes in a context where multiple other process instances are running concurrently, i.e. at the same time.} 
Let us consider a help desk ticket management process, in which two tickets may be in exactly the same process state, e.g., \bdq{assigned to an operator}. Their evolution may differ due to the context in which they are executed. In the first case, the ticket is handled quickly because only a few other tickets are open at that time and the operators have a low workload. In the second case, instead, the ticket experiences delays because many other urgent requests are being processed concurrently, occupying the same operators. 

The proposed study investigates whether modeling contextual process instances provides benefits for the next activity prediction task, i.e., a sub-field of next event prediction \cite{DiFrancescomarino2022}, where the objective is to predict the next activity to be executed given a running process instance. To this end, an approach based on the concept of Instance Graphs \cite{Diamantini2016} is proposed to model running process instances. Several encoding strategies are introduced to model contextual process instances, and their impact on Graph Neural Network performance is systematically evaluated.
More specifically, the proposed approach aims to address the following research questions:
\begin{itemize}
\item \textbf{RQ1}: which encoding strategy achieves the best prediction performance?
\item \textbf{RQ2}: how the best strategy performs compared with \rev{graph-based approaches for next activity prediction?}
\end{itemize}
The rest of this manuscript is structured as follows. \rev{Section \ref{sec:related} provides an overview of existing methods for predictive process monitoring, focusing on graph-based approaches and the modeling of inter-case dependencies}. Section \ref{sec:methodology} illustrates the proposed methodology. Section \ref{sec:experiments} illustrates the experimental settings, the dataset, and the results achieved. Finally, Section \ref{sec:conclusions} concludes the paper and delineates future research works.

\section{Related work}
\label{sec:related}
\subsection{Graph-based Predictive Process Monitoring}
\label{sec:related_graph}
\rev{Graph-based approaches represent a promising direction in PPM \cite{dissegna2024graph}, as graphs naturally represent process control-flow and can capture complex behaviors, such as parallelism and non-linear dependencies. 
Among existing graph-based approaches, \cite{chiorrini2023multi} leverage Instance Graphs and a Deep Graph Convolutional Neural Network for next activity prediction, incorporating control-flow and multiple temporal perspectives. \cite{intercase} extend this approach by incorporating contextual features, namely the number of active cases, running activities, and resource workload. \cite{prophet} represent cases as heterogeneous graphs with activity, resource, time, and trace nodes, and employ Graph Attention Networks (GATs) to predict the next activity. Similarly, \cite{dissegna2025multi} model traces as heterogeneous graphs and use GATs to predict the complete set of attributes of the next event. For object-centric event logs, \cite{adams2023preserving} preserve the graph-based representation by explicitly modeling interactions among different objects, whereas the proposed approach relies exclusively on case, activity, and timestamp attributes, avoiding domain-specific adaptations. Finally, \cite{ruffini2026leveraging} introduce a status graph that represents the history of ongoing executions for next activity and remaining time prediction. Unlike the proposed approach, their method does not explicitly model temporal dependencies and represents contextual process instances as disconnected components, limiting the message-passing capabilities of Graph Neural Networks. Moreover, the impact of different strategies for incorporating contextual instances is not investigated.}

\subsection{Inter-case Dependencies in Predictive Process Monitoring}
\label{sec:related_intercase}
\rev{Several studies incorporate inter-case information into PPM by engineering features from concurrently running cases.
\cite{SENDEROVICH2019255} incorporate inter-case dependencies by encoding features from concurrently running cases through knowledge-driven and data-driven approaches. \cite{9229927} introduce inter-case features for remaining time prediction to capture dynamics such as batching and improve predictive performance. \cite{GUNNARSSON2024102432} propose the LS-ICE framework to encode the state of relevant process load points as inter-case features for remaining trace and runtime prediction. \cite{10.1007/978-3-032-13426-4_4} propose I3SP, an encoder-decoder architecture that directly learns inter-case dependencies from the log prefix for suffix prediction, without manually engineered inter-case features. 
These works demonstrate the potential of inter-case information to improve predictive performance, but the explicit modeling of structural relationships between events belonging to concurrently running cases, particularly for next activity prediction, remains less explored. The proposed paper addresses this gap by investigating different graph-based strategies for encoding contextual events and evaluating their impact on next activity prediction.
}

\section{Methodology}
\label{sec:methodology}
The starting point is an \textit{event log}, consisting of \textit{traces} tracking process executions (or \textit{cases}). Each trace $\sigma=\langle e_1, \ldots, e_n \rangle$ consists of a finite sequence of \textit{events}. Each event corresponds to the execution of a process activity and is described by at least one timestamp and the name of the corresponding activity. The second input is a process model, mined using the Inductive Miner algorithm \cite{leemans2013discovering} by varying the noise threshold to obtain a model with a precision of at least 80\%. The first step of the methodology is \textit{Building Instance Graphs}, which converts log traces in the corresponding Instance Graphs (IGs). The set of the IGs is then the input for \rev{the} \textit{Feature engineering} step, which is responsible for enriching the IGs with temporal and contextual features. 
Following, in the \textit{Data Encoding} step, the set of prefix-IGs is derived, and several encoding strategies to incorporate contextual process instances are proposed. Each resulting representation is then provided as input to \textit{Graph Neural Network} to perform the prediction.

\subsection{Building Instance Graphs}
\begin{figure*}[tb]
\centering
\subfloat[]{
\includegraphics[width=0.5\textwidth]{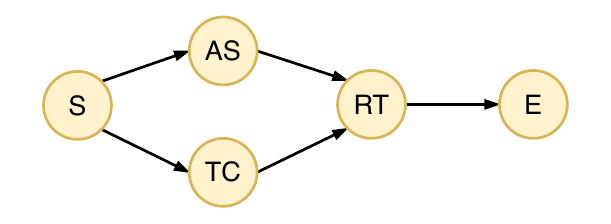}
\label{fig:ig}}
\hfil
\subfloat[]{
\includegraphics[width=0.25\textwidth]{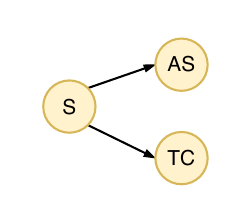}
\label{fig:prefix_ig}}
\caption{The IG derived from Table \ref{table:event_log} (a) and its prefix-IG of size 3 (b).}
\end{figure*}

An Instance Graph (IG) is a directed, acyclic graph that describes a specific execution of a process. Given a trace $\sigma$, its IG is defined as $\gamma_{\sigma}=(E, W)$, where each node in the set $E$ corresponds to an event in $\sigma$ and the set of edges $W$ models \textit{Causal Relations} (CRs) between process activities. Informally, a CR between activities $A$ and $B$ denotes that the execution of $B$ depends on the execution of $A$. Consequently, IGs explicitly model parallelism, i.e., lack of CRs, among process activities. Table \ref{table:event_log} shows an excerpt of the Helpdesk event log, while Figure \ref{fig:ig} shows the corresponding IG, with activities represented by acronyms.

\begin{table}[tb]
\centering
\begin{tabular}{l l l l}
\toprule
\textbf{Case ID} & \textbf{Event ID} & \textbf{Activity} & \textbf{Timestamp} \\
\midrule
\multirow{5}{*}{1} & 1 & Start & 2012-04-03 16:55 \\
& 2 & Assign Seriousness & 2012-04-03 16:55 \\
& 3 & Take in Charge ticket & 2012-04-03 16:55 \\
& 4 & Resolve Ticket & 2012-04-05 17:15 \\
& 5 & End & 2012-04-05 17:15 \\
\bottomrule
\end{tabular}
\caption{A trace from the Helpdesk event log.}
\label{table:event_log}
\end{table}

\subsection{Feature engineering}
\label{sec:features_eng}
This section enrich\rev{es} the set of IGs with multiple perspectives, capturing temporal and contextual aspects. \rev{To this end, each event must have explicit start and end timestamps; if one is missing, the procedure described in \cite{intercase} is applied to make both timestamps explicit.}

\subsubsection{Temporal enrichment}
The data payload associated with the events is leveraged to define multiple temporal features \cite{chiorrini2023multi}. 
Let $\sigma=\langle e_1, \ldots, e_n \rangle$ be a trace. The first temporal feature defined is $\Delta_{t_{e_i}}$, which represents the time between the current event and its predecessor. \rev{In addition, $t_{d_{e_i}}$ represents the time at which the event occurred with respect to the start of the process, while $t_{w_{e_i}}$ captures the point in the corresponding working week at which the event occurred, i.e, the elapsed time since midnight on the previous Sunday.}

\subsubsection{Contextual enrichment}
\begin{figure*}[tb]
\centering
\subfloat[]{
\includegraphics[width=0.5\textwidth]{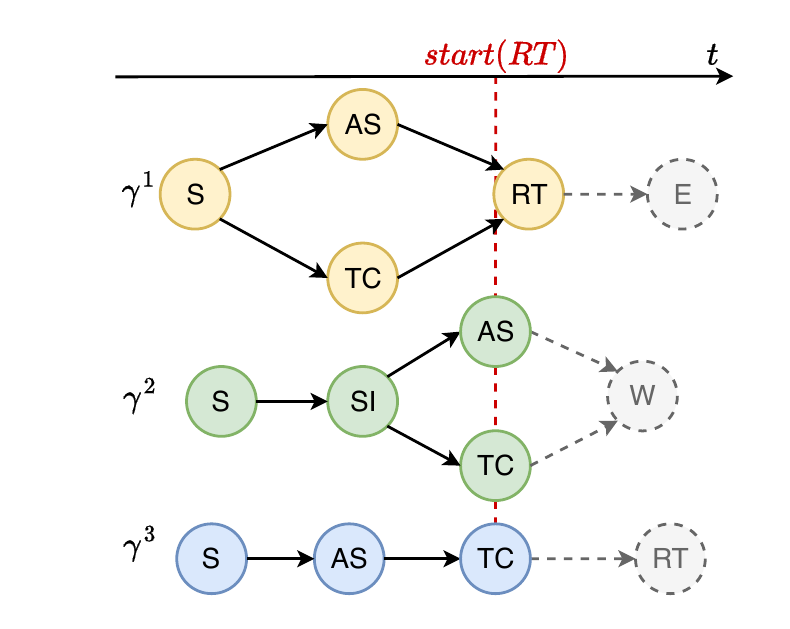}
\label{}}
\hfil
\subfloat[]{
\includegraphics[width=0.4\textwidth]{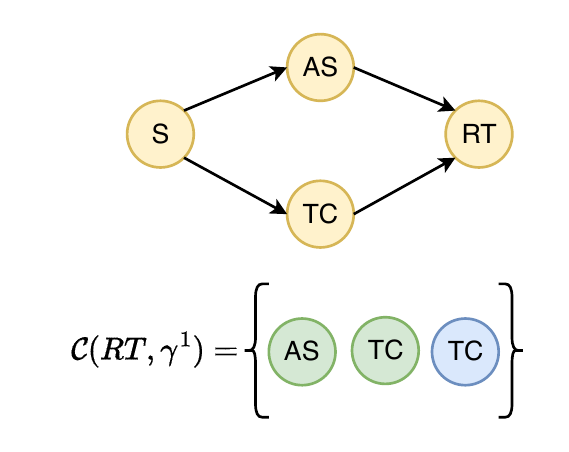}
\label{}}
\caption{An example of scenario in which $\gamma^1$ is the case under analysis (a), and $RT$ the event for which the \emph{set of contextual events} is identified (b).}
\label{fig:concurrent_events}
\end{figure*}

To capture contextual perspectives, the proposed approach draws inspiration from \cite{intercase}, where multiple aggregated features are derived from contextual process executions. \rev{Let $\mathcal{G}$ be the set of IGs,} $\gamma=(E,W) \in \mathcal{G}$ be the IG of a trace $\sigma$, $e_i \in E$ an event of $\gamma$, and $act(e_i)$, $start(e_i)$, $end(e_i)$ be respectively the activity, the start and the end timestamps of $e_i$. The \textit{set of contextual events} of $e_i$ is defined as $\mathcal{C}(e_i,\gamma)=\{\langle e_j, \gamma' \rangle \mid \gamma'=(E',W') \in \mathcal{G} \land \gamma'\neq \gamma \land e_j \in E' \land  start(e_j)\leq start(e_i)\leq end(e_j)\}$. Informally, $\mathcal{C}(e_i,\gamma)$ represents the events from the contextual process instances that are in execution when $e_i$ of $\gamma$ is running.
Let Figure \ref{fig:concurrent_events} illustrates an example scenario in which $\gamma^1$, $\gamma^2$, and $\gamma^3$ represent three running cases. Let $\gamma^1$ and $RT$ denote, respectively, the case under analysis and the target event. The resulting set of contextual events is $\mathcal{C}(RT, \gamma^1)=\{\langle AS, \gamma^2 \rangle, \langle TC, \gamma^2 \rangle, \langle TC, \gamma^3 \rangle\}$.

\subsection{Data encoding}
\label{sec:data_encoding}
In order to build a model capable of making prediction at each stage of a process execution, the set of partial process executions, i.e., the set of prefix-IGs, is derived. Let $\gamma \in \mathcal{G}$ be an IG. Its prefix-IG of size $k$ ($p_k(\gamma)$) represents the subgraph of $\gamma$ composed by the first $k$ events (nodes), and the corresponding label, i.e., the next activity, is associated to the activity of the event in position $k+1$. Figure \ref{fig:prefix_ig} displays an example of prefix-IG, where the label corresponds to the activity $RT$.

To incorporate the \emph{set of contextual events}, i.e., the set $\mathcal{C}(e_k,\gamma)$ of events in execution when $p_k(\gamma)$ is running, different encoding strategies are proposed. The first two strategies model a single graph representing both the prefix-IG and its set of contextual events. The remaining strategies model a first graph representing the prefix-IG, and a \emph{context graph} modeling its contextual events. Figure \ref{fig:concurrent_events} is used as example scenario.

\subsubsection{Encoding strategy 1 (E1)}
This strategy augments the prefix-IG by adding, for each $\langle e_j, \gamma' \rangle \in \mathcal{C}(e_k,\gamma)$, a node representing the corresponding contextual event together with an edge originating from the first node of $p_k(\gamma)$, and pointing to the newly added node. In this way, contextual events are represented as \rev{starting simultaneously with} the current prefix-IG, as illustrated in Figure \ref{fig:encoding_var_1}. \dop{In that case, the contextual events are as if they were constrained only to the execution of the first event of $p_k(\gamma)$.}

\subsubsection{Encoding strategy 2 (E2)}
This strategy is similar to the previous one, with the difference that contextual events are represented \rev{as starting simultaneously with} the last node of $p_k(\gamma)$, namely $e_k$. Accordingly, edges are added from the nodes having an outgoing edge toward $e_k$, to the corresponding contextual nodes, as illustrated in Figure \ref{fig:encoding_var_2}. \dop{This encoding is more constrained than that of E1, since all contextual events are effectively assumed to start simultaneously with $e_k$.}

\subsubsection{Encoding strategy 3 (E3)}
This strategy models two distinct graphs. The first represents $p_k(\gamma)$, while the second corresponds to a \emph{context graph} containing all events in $\mathcal{C}(e_k,\gamma)$. To ensure the correct functioning of the message-passing procedure (Section \ref{sec:gnn}), the \emph{context graph} should be defined as a connected component. To this end, the first node of $p_k(\gamma)$ is duplicated within the \emph{context graph}, together with edges originating from this node and pointing to each contextual node, as illustrated in Figure \ref{fig:encoding_var_3}. \dop{Encoding E3 is very similar to E1, except that the execution of the \emph{context graph} is independent of the execution of $p_k(\gamma)$.}

\subsubsection{Encoding strategy 4 (E4)}
\begin{figure*}[tb]
\centering
\subfloat[E1.]
{\includegraphics[width=0.24\textwidth]{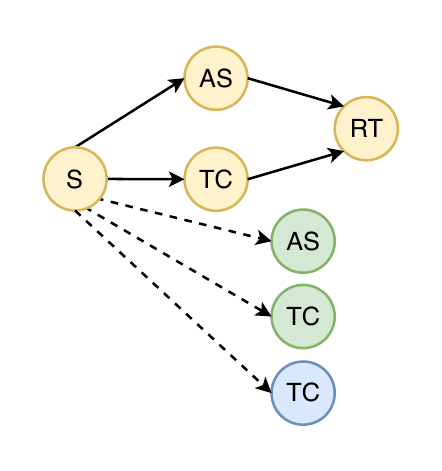}
\label{fig:encoding_var_1}}
\subfloat[E2.]
{\includegraphics[width=0.24\textwidth]{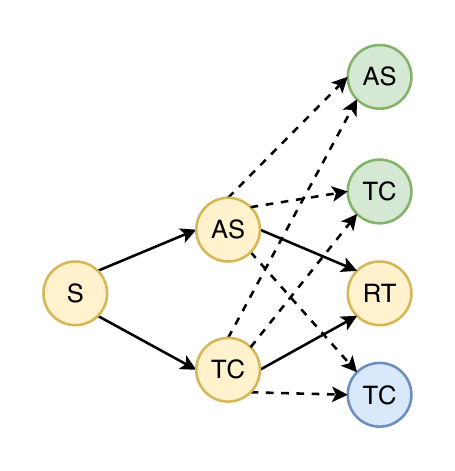}
\label{fig:encoding_var_2}}
\subfloat[E3.]
{\includegraphics[width=0.24\textwidth]{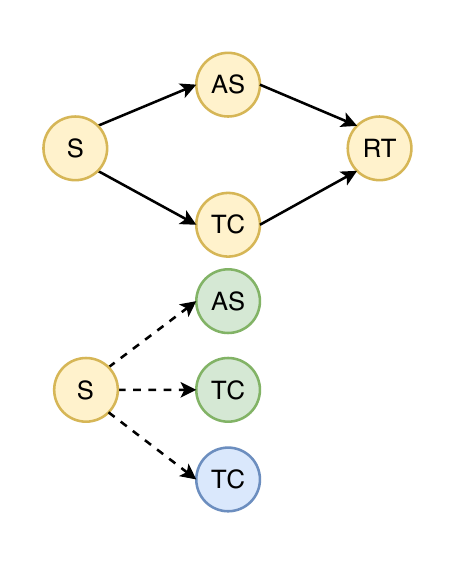}
\label{fig:encoding_var_3}}
\subfloat[E4.]
{\includegraphics[width=0.24\textwidth]{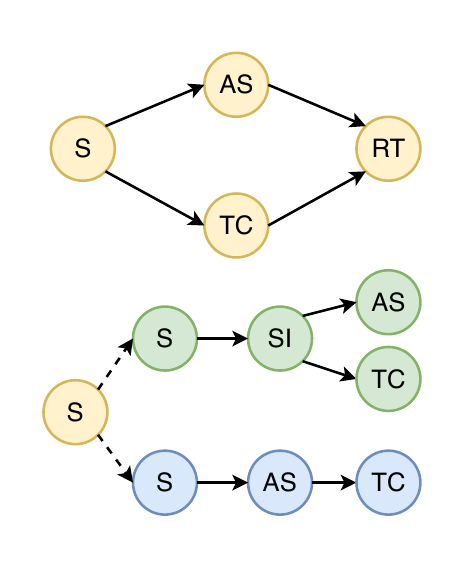}
\label{fig:encoding_var_4}}
\caption{The proposed encoding strategies.}
\label{fig:encoding}
\end{figure*}

The last strategy augments the \emph{context graph} by incorporating the complete contextual prefix-IGs. More specifically, for each $\langle e_j,\gamma' \rangle \in \mathcal{C}(e_k,\gamma)$, the prefix-IG $p_j(\gamma')$ is added to the \emph{context graph}, together with an edge originating from the first event of $p_k(\gamma)$ and pointing to the first event of $p_j(\gamma')$. When multiple contextual events belong to the same case, only the prefix-IG which includes all the others is kept (Figure \ref{fig:encoding_var_4}). 

This strategy is similar to the approach proposed in \cite{ruffini2026leveraging}, with two main differences: (i) temporal features are explicitly modeled, and (ii) the \emph{context graph} is represented as a single connected component.

\subsection{Graph Neural Network}
\label{sec:gnn}

To perform prediction, the proposed approach is based on a spatial-based Graph Convolutional Neural Network \cite{Wu2021}, where each convolutional layer implements a message-passing step, enabling nodes to exchange information with their neighbors. Let $G=(E, W)$ be a graph, $u,v \in E$ be two nodes of $G$, and $(u,v)$ be the edge of $G$ that connects nodes $u$ and $v$. The graph $G$ is described by the feature matrix ($X \in \mathbb{R}^{|E| \times c}$), where each node is encoded using the one-hot representation of its corresponding activity and three temporal features (Section \ref{sec:features_eng}), and the adjacency matrix ($A \in \mathbb{R}^{|E| \times|E|}$), that encodes graph topology. To compute node embeddings, the formulation proposed in \cite{10.5555/3294771.3294869} is adopted. Specifically, for each $u \in E$, $h_u^{'} = \sigma \left(W^{self}h_u + \frac{1}{\mid \mathcal{N}^*\left(u\right)\mid} \sum_{v \in \mathcal{N}^*\left(u\right)}{W^{node}h_v}\right)$, where $\sigma$ is the Rectified Linear Unit function, $W^{self}$ and $W^{node}$ are matrices with learnable parameters, and $\mathcal{N}^*\left(u\right)$ is a random sample of $u$'s neighbors. 
Since each graph is associated with a label, the problem is formulated as a graph classification. 

\begin{figure}[tb]
\centering
\includegraphics[width=\textwidth]{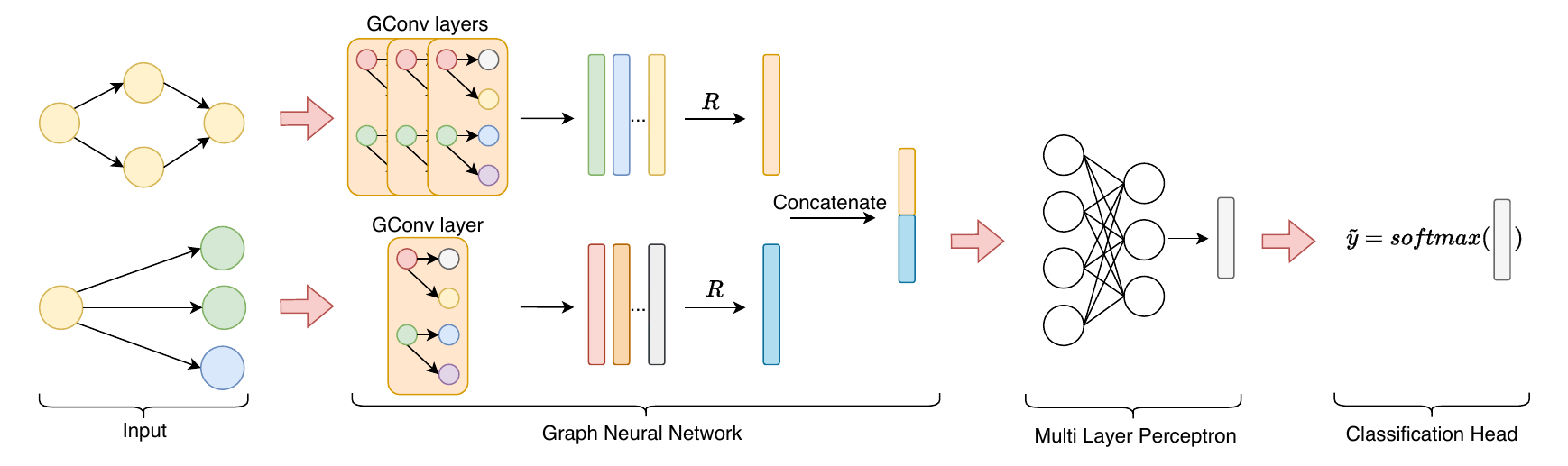}
\caption{The adopted network architecture for E3 and E4.}
\label{fig:gnn}
\end{figure}

A readout function ($R$) generates a global graph representation, which is passed to an Multi Layer Perceptron (MLP) and a Softmax classification layer. Two architectural variants are further designed: (i) for a single graph, $K$ convolutional layers are followed by $R$, the MLP, and the classification head, (ii) for separate graphs, the prefix-IG is processed by $K$ convolutional layers, while the \emph{context graph} uses a single convolutional layer to capture high-level contextual information. The resulting graph embeddings are obtained through separate $R$ functions, concatenated, and fed into the MLP and classification head. Figure \ref{fig:gnn} illustrates the adopted network architecture for the strategies in which two separated graphs are derived.

\section{Experiments}
\label{sec:experiments}
To address both the research questions, for each dataset, log traces are chronologically sorted from the earliest to the most recent, reserving the first 67\% of the cases for the training set and the remaining 33\% for the test set, \rev{with} the last 20\% of the training set \rev{used for} validation. 
\dop{A hyperparameter tuning procedure for the GNN is carried out using \href{https://optuna.org}{Optuna}. The search space \rev{includes} a learning rate in $[10^{-2},10^{-5}]$, the number of hidden graph convolutional layers $K\in [0, 4]$, the size of the convolutional layers in $[64, 256]$, the dropout in $[0.0, 0.4]$, and the readout function $R\in\{add,mean,max\}$.}
Each configuration \rev{is} trained for 200 epochs, with an early stopping on the validation loss to prevent overfitting. \rev{To address RQ1, hyperparameter search is performed for each encoding strategy, and the results are compared. To address RQ2, the best-performing encoding strategy, identified as the one achieving the lowest validation loss on the largest number of datasets, is compared against multiple graph-based approaches for the next activity prediction \cite{chiorrini2023multi,intercase,prophet}.} To prevent data leakage, the process model is discovered from the training set and then used to construct IGs \rev{for all} traces; feature normalization is computed using the maximum values observed in the training data; for each prefix, \rev{the corresponding contextual events are always taken from the same data split as the prefix}. \rev{This means that, if a prefix belonging to the training set has contextual events referring to process instances assigned to the validation or test sets, those events are excluded. It is worth noting that this choice may lead to an asymmetric and potentially sparse contextual representation near the train-test boundary, a phenomenon that does not affect the experimental comparisons presented below and whose impact on performance is left for future investigation.}

To ensure a fair comparison, the experiments of the competitors are repeated \rev{using their original hyperparameter search spaces,} adopting the \rev{case} split, prefix generation criteria, and evaluation metrics proposed in this paper. 
\rev{For each dataset and for every encoding strategy and competing approach, the model was trained 5 times using the same hyperparameter configuration but different random seeds.
The reported Accuracy (Acc), Macro F1-score (F1-M), and Weighted F1-score (F1-W) are the mean and standard deviation over the resulting five independent runs.}
The source code \rev{and details of experiments} are available at the following GitHub repository\footnote{\rev{https://github.com/AlessandroMele/leveraging\_contextual\_events\_nap}}.

\subsubsection{Dataset}
\label{sec:dataset}
To evaluate the proposed approach, various real-world event logs publicly available\footnote{https://data.4tu.nl} are used. The B\rev{PI} Challenge 2012 dataset contains events from a loan application process. The log consists of different subprocesses, and the completed variant of the W-subprocess (BPI12WC) is selected. The BPI challenge 2020 dataset contains events of travel expense claims. Among the different event logs, Prepaid Travel Cost (BPI20P) and Request for Payment (BPI20R) are selected. The Helpdesk dataset contains events from a ticketing management process of the help desk of an Italian software company. The Receipt dataset contains events produced in an anonymous municipality in the Netherlands during the receiving phase of the building permit application process. Table \ref{table:dataset_stats} illustrates an overview of the selected dataset.
\begin{table}[tb]
\centering
\caption{Event log statistics.}
\label{table:dataset_stats}
\setlength{\tabcolsep}{1pt}
\begin{tabular}{l c c c c c c c} 
\toprule
\multirow{2}{*}{\textbf{Dataset}} & \multirow{2}{*}{\textbf{N.traces}} & \multirow{2}{*}{\textbf{N.events}} & \multirow{2}{*}{\textbf{N.activities}} & \multicolumn{3}{c}{\textbf{Trace length}} \\
\cmidrule{5-7}
 &  &  & & \textbf{Min} & \textbf{Max} & \textbf{Avg} \\
\midrule
BPI12WC & 9658 & 72413 & 6 & 1 & 74 & 7.4 \\
BPI20P & 2099 & 18246 & 29 & 1 & 21 & 8.6 \\
BPI20R & 6886 & 36796 & 19 & 1 & 20 & 5.3 \\ 
Helpdesk & 3804 & 13710 & 9 & 1 & 14 & 3.6 \\
Receipt & 1434 & 8577 & 27 & 1 & 25 & 5.9 \\
\bottomrule
\end{tabular}
\end{table}

\subsubsection{RQ1}
From Table \ref{table:encoding_results}, it can be observed that there is no single strategy that clearly dominates the others. Strategy E2 performs the worst, showing substantial negative gaps in \rev{all the metrics} with respect to the best performing strategy \rev{(except on F1-M for Receipt).} 
Conversely, \rev{the E1 strategy consistently outperforms E2, with an average improvement of 4.95\% in Acc, 2.89\% in F1-M and 4.58\% in F1-W.} 
Strategy E3 is almost always the second best approach \rev{in Acc (except for BPI20P, BPI20R), and F1-W (except for BPI20P)}. Finally, the best-performing strategy is E4, where the \emph{context graph} tracks the entire history of the contextual events\rev{, although it provides only a negligible improvement over E3.} \rev{Since E4 is the encoding strategy achieving the lowest validation loss on three out of the five datasets, it is selected for comparison with the competing approaches.}

\begin{table}[tb]
\centering
\caption{\rev{Mean and standard deviation (in brackets) of the test set results for different encoding strategies}. The best results are shown in bold, while the second-best are underlined.}
\label{table:encoding_results}
\setlength{\tabcolsep}{1.5pt}
\resizebox{\textwidth}{!}{%
\begin{tabular}{l l c c c c c}
\toprule
\textbf{Encod.} & \multirow{2}{*}{\textbf{Metrics}} & \multicolumn{5}{c}{\textbf{Dataset}} \\
\cmidrule{3-7}
\textbf{Strat.} & & \textbf{BPI12WC} & \textbf{BPI20P} & \textbf{BPI20R} & \textbf{Helpdesk} & \textbf{Receipt} \\
\midrule

\multirow{3}{*}{E1} & Acc & 77.16(0.14) & 85.37(0.24) & \underline{87.04(0.07)} &  78.97(1.03) & 85.14(0.39) \\
& F1-M & 53.95(1.35) & 48.10(0.30) & \underline{41.27(0.84)} & \underline{28.83(2.79)} & \underline{39.18(4.79)} \\
& F1-W & 73.37(0.63) & \textbf{83.42(0.26)} & 82.13(0.14) & 73.69(1.56) & 83.12(0.52)\\
\midrule

\multirow{3}{*}{E2} & Acc & 71.16(0.32) & 78.05(0.68) & 83.89(0.52) & 75.51(1.47) & 80.32(0.21) \\
& F1-M & 51.29(1.09) & 38.26(1.89) & 38.84(1.22) & 26.85(3.48) & \textbf{41.62(0.77)} \\
& F1-W & 68.34(0.53) & 76.00(0.73) & 79.67(0.61) & 70.39(2.11) & 78.43(0.19) \\
\midrule

\multirow{3}{*}{E3} & Acc & \underline{77.45(0.04)} & \textbf{85.46(0.12)} & 87.00(0.15) & \underline{80.31(0.07)} & \underline{85.43(0.24)} \\
& F1-M & \underline{60.99(0.26)} & \underline{48.15(0.27)} & \textbf{42.72(0.41)} & 30.23(0.02) & 40.98(4.04) \\
& F1-W & \underline{74.25(0.14)} & 82.82(0.11) & \underline{82.35(0.06)} & \underline{75.12(0.06)} & \underline{83.39(0.30)} \\
\midrule

\multirow{3}{*}{E4} & Acc & \textbf{77.46(0.05)} & \underline{85.41(0.46)} & \textbf{87.52(0.05)} & \textbf{80.77(0.16)} & \textbf{85.63(0.13)} \\
& F1-M & \textbf{61.20(0.26)} & \textbf{48.52(0.56)} & 40.83(1.22) & \textbf{30.48(0.14)} & \underline{41.13(2.45)} \\
& F1-W & \textbf{74.27(0.23)} & \underline{83.00(0.13)} & \textbf{82.72(0.10)} & \textbf{75.63(0.12)} & \textbf{83.66(0.17)} \\

\bottomrule
\end{tabular}
}
\end{table}

\subsubsection{RQ2}
From Table \ref{table:results}, it can be observed that \dop{the selected encoding strategy} \rev{enhances} performance, outperforming \rev{the selected graph-based} approaches. In details, the proposed approach outperforms both the baseline \cite{chiorrini2023multi} and the approach based on aggregated contextual features \cite{intercase}. Compared with \cite{intercase}, \rev{an average improvement of 2.70\% in Acc, 4.57\% in F1-M, and 3.16\% in F1-W can be observed}. \rev{Compared with \cite{prophet}, which model log traces as heterogeneous graphs, the proposed approach achieves larger performance gains, with an average improvement of 7.75\% in Acc, 11.76\% in F1-M, and 7.90\% in F1-W}.

\begin{table}[tb]
\centering
\caption{\rev{Mean and standard deviation of the test set results}. The best results are shown in bold, while the second-best are underlined.}
\label{table:results}
\setlength{\tabcolsep}{1.5pt}
\resizebox{\textwidth}{!}{
\begin{tabular}{l l c c c c c}
\toprule
\multirow{2}{1.3cm}{\textbf{Approach}} & \multirow{2}{1.3cm}{\textbf{Metrics}} & \multicolumn{5}{c}{\textbf{Dataset}}\\
\cmidrule{3-7}
& & \textbf{BPI12WC} & \textbf{BPI20P} & \textbf{BPI20R} & \textbf{Helpdesk} & \textbf{Receipt} \\
\midrule

\multirow{3}{1.6cm}{Strategy E4} & Acc & \textbf{77.46(0.05)} & \textbf{85.41(0.46)} & \textbf{87.52(0.05)} & \textbf{80.77(0.16)} & \textbf{85.63(0.13)} \\
& F1-M & \textbf{61.20(0.26)} & \textbf{48.52(0.56)} & \textbf{40.83(1.22)} & \textbf{30.48(0.14)} & \textbf{41.13(2.45)} \\
& F1-W & \textbf{74.27(0.23)} & \textbf{83.00(0.13)} & \textbf{82.72(0.10)} & \textbf{75.63(0.12)} & \textbf{83.66(0.17)} \\
\midrule

\multirow{3}{*}{\cite{intercase}} & Acc & \underline{76.56 (1.62)} & \underline{81.96(1.19)} & 85.21(0.07) & \underline{78.31(0.40)} & 81.25(0.90) \\
& F1-M & \underline{58.25(1.61)} & 36.97(4.09) & 38.67(0.96) & \underline{29.61(0.13)} & \underline{35.77(2.40)} \\
& F1-W & \underline{72.43(2.17)} & \underline{78.18(1.62)} & 80.30(0.12) & \underline{73.31(0.36)} & 79.23(0.85) \\ 
\midrule

\multirow{3}{*}{\cite{prophet}} & Acc & 69.79(0.25) & 71.79(0.15) & 77.06(0.10) & 75.68(0.12) & \underline{83.72(0.44)} \\

& F1-M & 47.41(0.78) & 29.19(0.64) & 30.60(0.24) & 22.81(0.04) & 33.33(1.15) \\

& F1-W & 66.91(0.10) & 67.21(0.21) & 71.63(0.22) & 71.16(0.13) & \underline{82.87(0.47)} \\
\midrule

\multirow{3}{*}{\cite{chiorrini2023multi}} & Acc & 75.96(2.11) & 80.12(6.38) & \underline{85.29(0.01)} & 78.09(0.57) & 83.22(0.62) \\
& F1-M & 52.67(8.44) & \underline{39.52(3.19)} & \underline{39.63(1.21)} & 28.92(1.58) & 34.59(1.04) \\
& F1-W & 71.90(2.88) & 76.75(6.40) & \underline{80.48(0.11)} & 72.96(0.84) & 81.15(0.66) \\

\bottomrule
\end{tabular}
}
\end{table}

\begin{figure*}[tb]
\centering
\subfloat[BPI12WC]
{\includegraphics[width=0.55\textwidth]{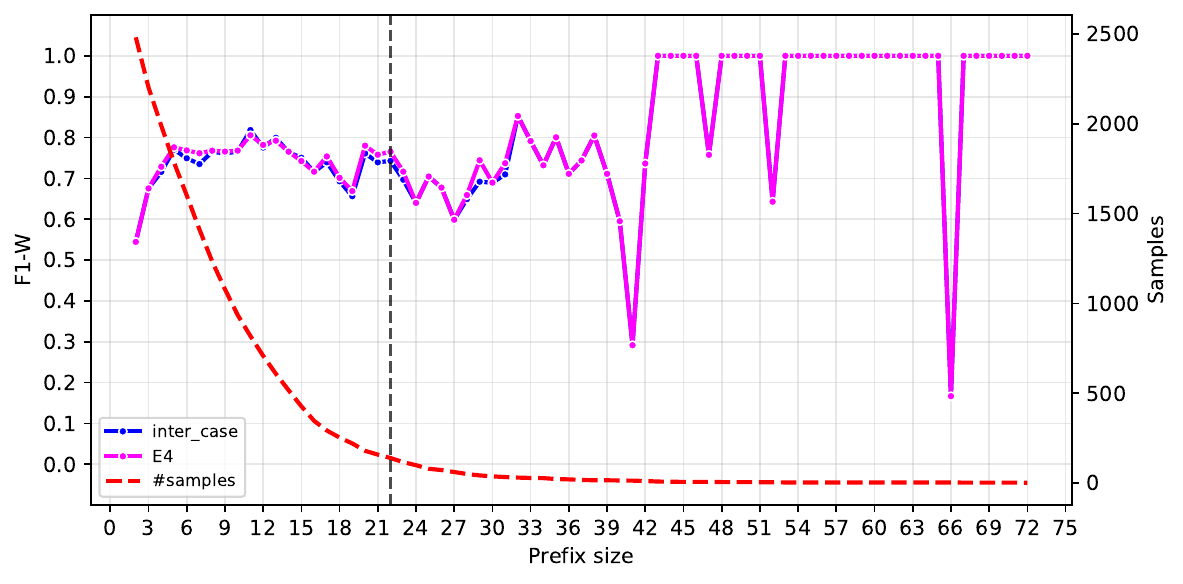}
\label{fig:prefix_metrics_test_BPI12WC}}
\hfil
\subfloat[BPI20P]
{\includegraphics[width=0.45\textwidth]{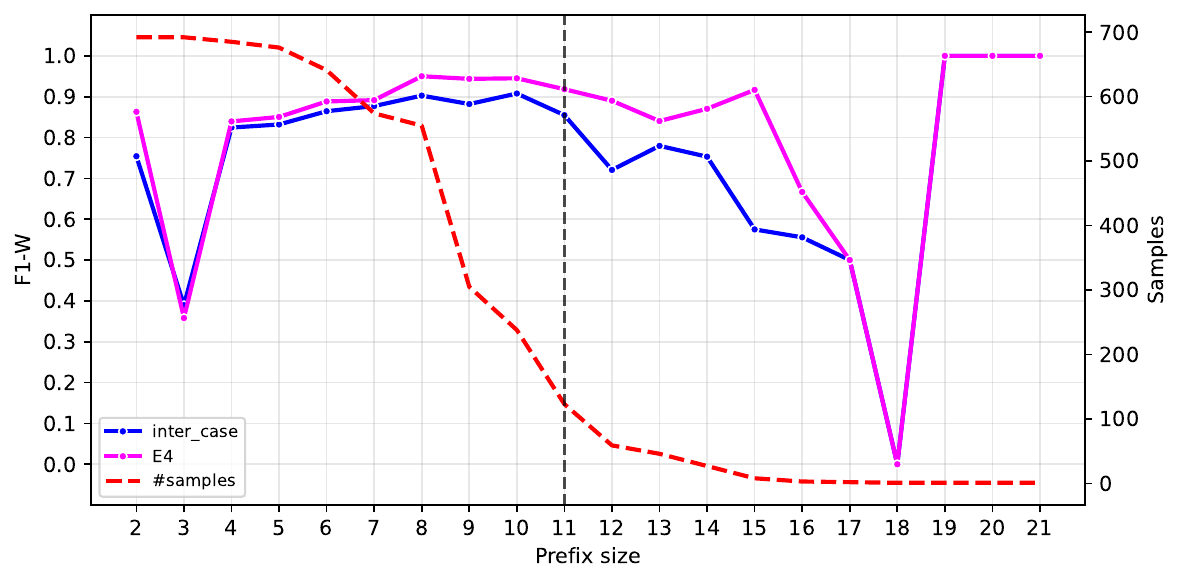}
\label{fig:prefix_metrics_test_BPI20P}}
\subfloat[BPI20R]
{\includegraphics[width=0.45\textwidth]{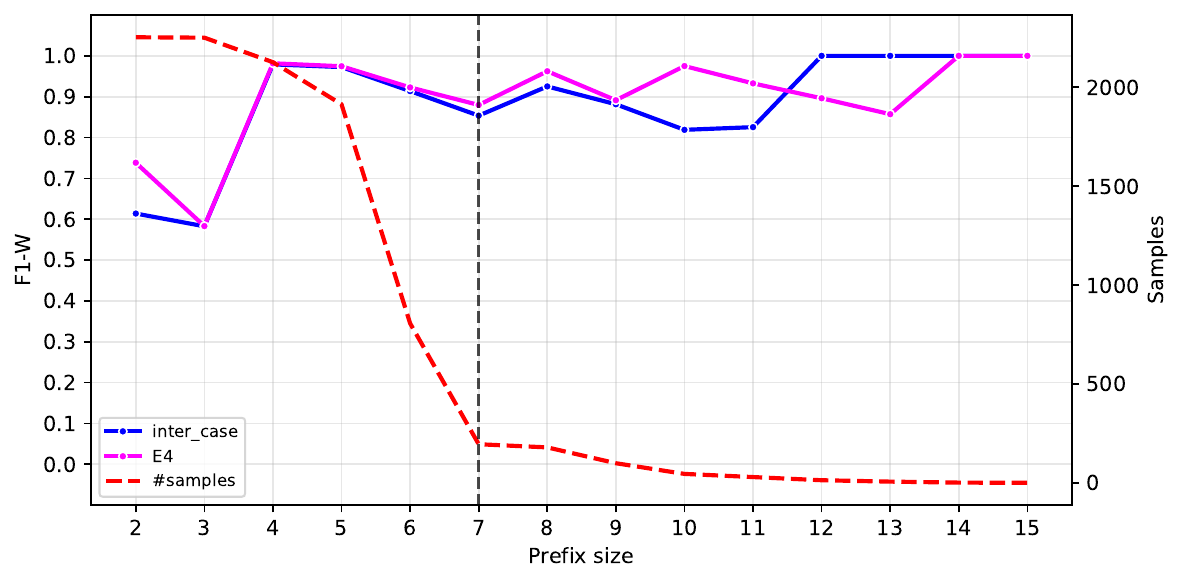}
\label{fig:prefix_metrics_test_BPI20R}}
\hfil
\subfloat[Helpdesk]
{\includegraphics[width=0.45\textwidth]{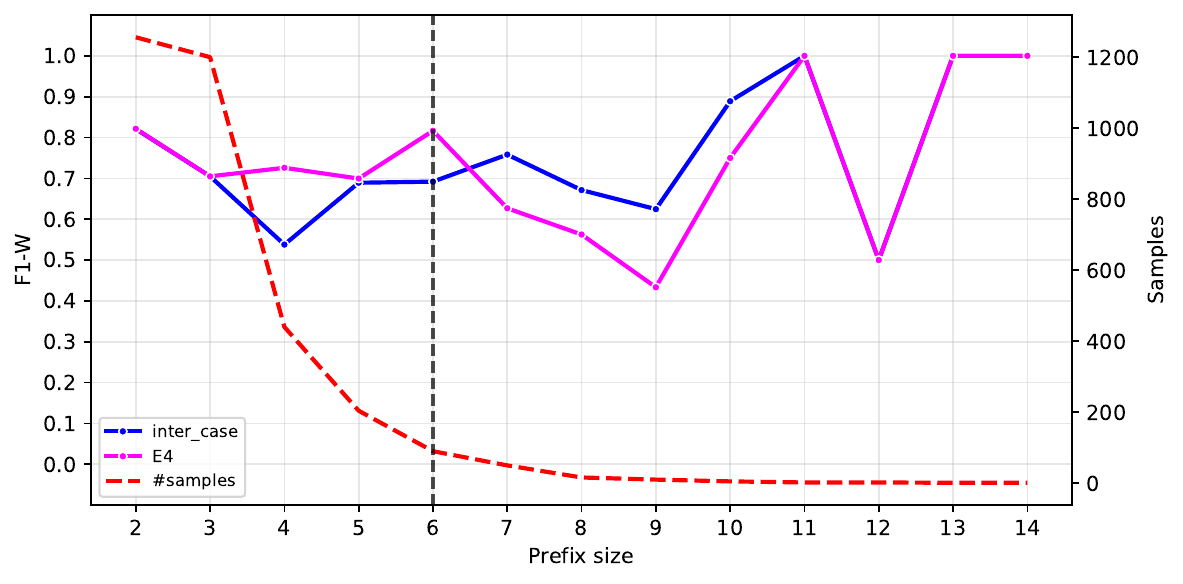}
\label{fig:prefix_metrics_test_Helpdesk_no_resources}}
\subfloat[Receipt]
{\includegraphics[width=0.45\textwidth]{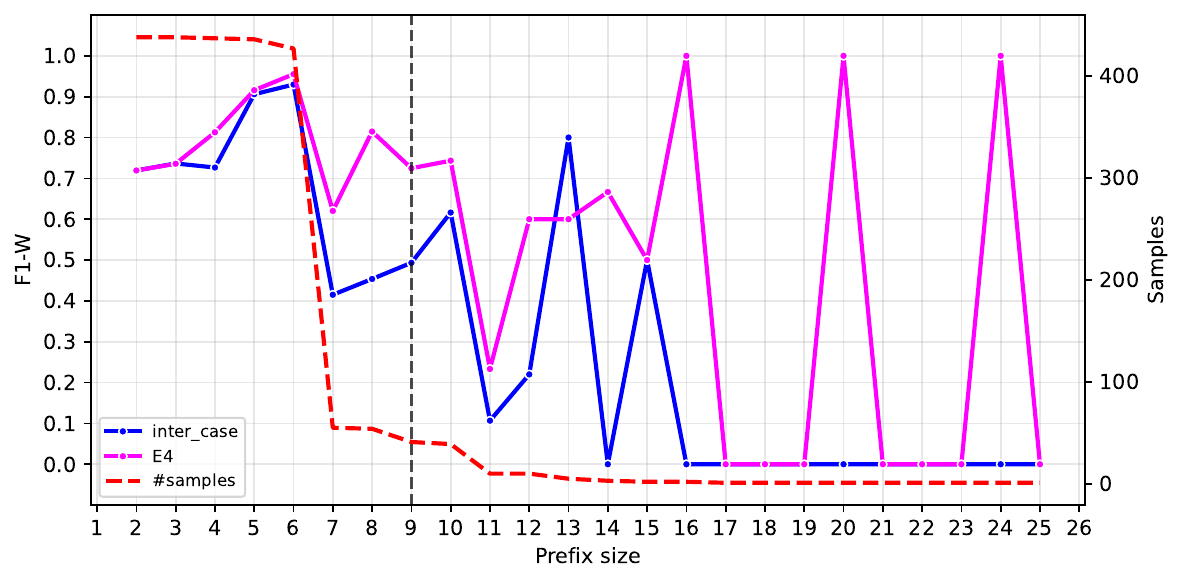}
\label{fig:prefix_metrics_test_Receipt}}
\caption{F1-W on test set varying the prefix-IG size.}
\label{fig:results_prefix_size}
\end{figure*}

More specifically, Figure \ref{fig:results_prefix_size} shows the F1-W metric with respect to the prefix-IG size, comparing the best-performing strategy with \cite{intercase}, which ranks as the second-best approach \rev{in the majority of cases}. Since the dataset contains significantly fewer large prefix-IGs, the F1-W becomes increasingly unstable for these sizes. For this reason, a vertical dotted line is used to indicate the point beyond which only 5\% of the samples remain in the dataset. Excluding these regions, the proposed approach achieves performance equal to or better than \cite{intercase}. 

\section{Conclusions and future work}
\label{sec:conclusions}
This paper presents a graph-based approach for the next activity prediction, modeling both process executions and contextual process instances. Multiple encoding strategies are introduced and evaluated to assess how different ways of incorporating contextual processes impact prediction performance. Experimental results on multiple event logs show that no single strategy consistently dominates across all datasets. However, the best overall performance is achieved by the strategies that model prefix-IGs and contextual process instances as two separate graphs, suggesting that explicitly separating process history from contextual information better captures contextual dependencies. Overall, the proposed approach demonstrates strong effectiveness, outperforming \rev{multiple} graph-based \rev{approaches}. \rev{The main limitation of the proposed approach is that contextual events are identified solely based on temporal overlap, without considering resources, activity types, or causal relationships among process instances. Addressing these aspects represents a promising direction for future research. Furthermore,} it would be valuable to investigate how prediction performance varies when introducing a threshold for selecting contextual events and/or cases.

\bibliographystyle{splncs04}
\bibliography{bibliography}

\end{document}